\documentclass[runningheads]{llncs}
\usepackage{graphicx}
\usepackage{tikz}
\usepackage{comment}
\usepackage{amsmath,amssymb} % define this before the line numbering.
\usepackage{color}
\usepackage[accsupp]{axessibility}  % Improves PDF readability for those with disabilities.

\usepackage{color}

\newcommand{\eg}{{\em e.g.}}
\newcommand{\ie}{{\em i.e.}}

\usepackage{booktabs}
\usepackage{multirow}

\begin{document}
% \linenumbers
\pagestyle{headings}
\mainmatter
\def\ECCVSubNumber{3700}  % Insert your submission number here
\newcommand{\repeatthanks}{\textsuperscript{\thefootnote}}
\title{FAMLP: A Frequency-Aware MLP-Like Architecture For Domain Generalization} % Replace with your title

\begin{comment}
\titlerunning{ECCV-22 submission ID \ECCVSubNumber} 
\authorrunning{ECCV-22 submission ID \ECCVSubNumber} 
\author{Anonymous ECCV submission}
\institute{Paper ID \ECCVSubNumber}
\end{comment}
% \begin{comment}
\titlerunning{Frequency-Aware MLP For Domain Generalization}
\author{Kecheng Zheng\dag \hspace{15pt}
Kai Zhu\dag\hspace{15pt}
Ruijing Zhao  \hspace{15pt}\\
Yang Cao \hspace{15pt}
Zheng-Jun Zha\thanks{Corresponding Author.} }

\authorrunning{K. Zheng et al.}
\institute{University of Science and Technology of China }
\renewcommand{\thefootnote}{\fnsymbol{footnote}}
\footnotetext[4]{Equal Contribution. }
% \footnotetext{$^\dagger$Equal contribution}
% \end{comment}
%******************
\maketitle
% \renewcommand{\thefootnote}{\fnsymbol{footnote}}
% \footnotetext[1]{Co-first Author}
% \footnotetext[2]{Corresponding Author}

\begin{abstract}
 MLP-like models built entirely upon multi-layer perceptrons have recently been revisited, exhibiting the comparable performance with transformers. It is one of most promising architectures due to the excellent trade-off between network capability and efficiency in the large-scale recognition tasks. However, its generalization performance to heterogeneous tasks is inferior to other architectures (\eg, CNNs and transformers) due to the extensive retention of domain information. To address this problem, we propose a novel frequency-aware MLP architecture, in which the domain-specific features are filtered out in the transformed frequency domain, augmenting the invariant descriptor for label prediction. Specifically, we design an adaptive Fourier filter layer, in which a learnable frequency filter is utilized to adjust the amplitude distribution by optimizing both the real and imaginary parts. A low-rank enhancement module is further proposed to rectify the filtered features by adding the low-frequency components from SVD decomposition. Finally, a momentum update strategy is utilized to stabilize the optimization to fluctuation of model parameters and inputs by the output distillation with weighted historical states. To our best knowledge, we are the first to propose a MLP-like backbone for domain generalization. Extensive experiments on three benchmarks demonstrate significant generalization performance, outperforming the state-of-the-art methods by a margin of $3\% $, $4\% $ and $9\% $, respectively.

\end{abstract}

\section{Introduction}

Convolutional neural networks (CNNs) brings huge performance breakthroughs to various vision tasks and dominates corresponding backbone networks (\eg, VGG, ResNet) for a long time. Recently, the transformer with self-attention mechanism replaces the local relation learning of CNNs with long-range modeling, pulling up the upper bound of the performance of deep networks. More recently, some MLP-like works further replace self-attention operations with only fully connected and skip-connected layers, achieving a better trade-off between network capability and efficiency.

Although MLP-like models show promising results in large-scale homogeneous recognition tasks (\eg, ImageNet classification), its transfer performance to various heterogeneous tasks is lower than that of CNNs and transformers with the same amount of parameters. To bridge this gap, this paper explores how MLP-like models trained from a collection of multiple training sources can be better generalized to unknown heterogeneous data domains, which is also known as the domain generalization (DG) problem. 

Existing DG works are mainly built upon CNNs~\cite{li2018learning,balaji2018metareg,xu2021fourier,carlucci2019domain} to learn domain-invariant representations after conditioning on the class label from known multi-source domains. They introduce adversarial training~\cite{li2018deep}, meta learning~\cite{balaji2018metareg}, self-supervised learning~\cite{xu2021fourier} or domain augmentation techniques~\cite{xu2021fourier} and have shown promising results. Orthogonally, some recent works extract generalized CNN features by augmenting the frequency domain, and find that manipulation on the amplitude components can directly affect the domain information. 

\begin{figure}[t!]
  \centering
  \includegraphics[width=.999\textwidth]{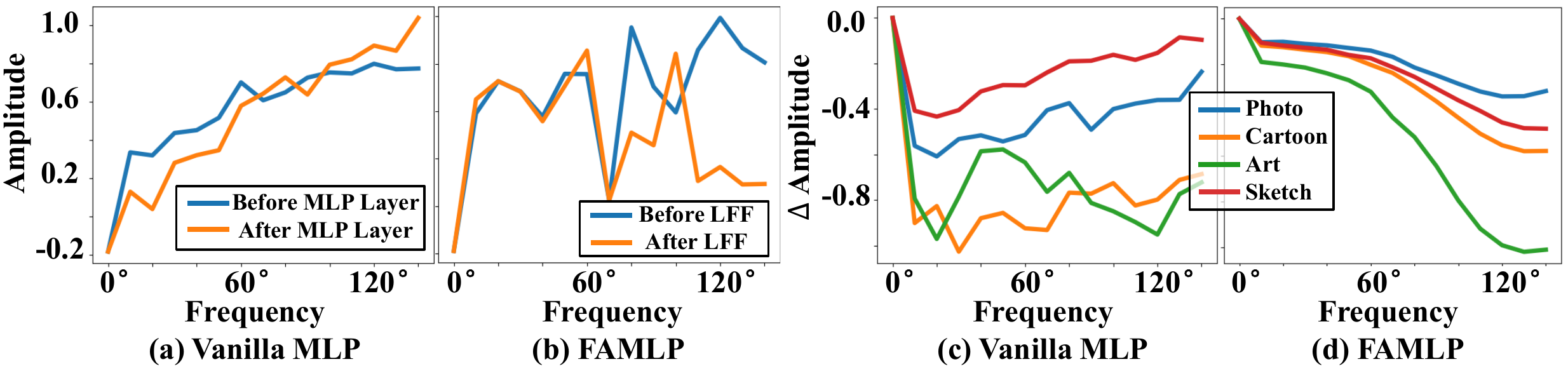}
  \caption{The Fourier analysis on PACS dataset. (a) The amplitudes before the last MLP layer and after it; (b) The amplitudes before the proposed adaptive Fourier filter and after it; (c) $\Delta$ amplitude is the difference between the amplitudes before the last MLP layer and after it on different domains; (d) $\Delta$ amplitude is the difference between the amplitudes before the proposed Fourier filter and after it on different domains.
  }
  \label{intro}
\end{figure}

Motivated by this, we analyze the SOTA MLP-like models in the frequency domain, revealing the following challenges. Firstly, as shown in Fig. \ref{intro} (a), we calculate the degree of filtering for different frequency components before/after MLP layer, which illustrates that MLP layer cannot suppress the high frequency components of the input features. When no extra frequency operations (\eg, Fourier filter in Fig. \ref{intro} (b)) are implemented, most high-frequency information are retained after the pure MLP layer, making it hard to resist the interference of heterogeneous data in domain generalization. Secondly, the frequency response is domain-specific. It can be seen in Fig. \ref{intro} (c) that the frequency responses are inconsistent between different domains, so it is impractical to set a fixed cutoff frequency. Meanwhile, the parameters of MLP-like models are data-independent, making it impossible to adjust the frequency response adaptively according to the input domain characteristics as shown in Fig. \ref{intro} (d). This makes MLP-Like architectures not suitable for DG problem, which covers the class prediction with different frequency distributions.

% Motivated by this, we analyze the SOTA MLP-like models in the frequency domain, revealing the following challenges. Firstly, MLP layer cannot suppress the high frequency components of the input features. As shown in Fig. \ref{intro} (a), we calculate the degree of filtering for different frequency components. When no extra frequency operations (\eg, Fourier filter in Fig. \ref{intro} (b)) are implemented, most high-frequency information is retained after the pure MLP layer, making it hard to resist the interference of heterogeneous data in domain generalization. Secondly, the frequency response is domain-specific. It can be seen in Fig. \ref{intro} (c) that the frequency response are inconsistent between different domains, so it is impractical to set a fixed 
% cutoff frequency. Meanwhile, the parameters of MLP-like models are data-independent, making it impossible to adjust the frequency response adaptively according to the input domain characteristics as shown in Fig. \ref{intro} (d). This makes MLP-Like architectures not suitable for DG problem, which covers the class prediction with different frequency distributions.

To address this problem, we propose a frequency-aware MLP framework (FAMLP), explicitly promoting the extraction of domain-invariant frequency features. The core of the framework is the adaptive Fourier filter layer, which enhances the rectification of low-frequency features block by block, mitigating the interference of domain shifts. Specifically, we firstly utilize the fast Fourier transform to map the features to the frequency domain before each MLP layer. Then the domain-specific features are filtered out by a learnable frequency filter, which corresponds to the real and imaginary parts of the frequency features. To ensure integrity of important features, the filtered features are further strengthened by fusing the low-frequency components from SVD decomposition. Finally, the domain-invariant features are mapped back to the spatial domain through the inverse Fourier transform for the subsequent MLP layer. Furthermore, to improve the overall generalization of the model from an optimization perspective, we propose a momentum update strategy, distilling the invariant features from a updated teacher model. We calculate the teacher model based on the weighted historical states of our FAMLP model, guaranteeing consistency of output for minor network changes. The input images obtained by data augmentation are fed into the teacher network to guide the optimization process in terms of robustness to different domain shifts.

The main contributions of this paper are three-fold:

\begin{itemize} 
    \item We propose a frequency-aware MLP framework (FAMLP) for domain generalization task, in which the low-frequency features are adaptively enhanced by a learnable frequency kernel, resulting in a domain-invariant representation. 
    \item We propose a momentum update strategy for the FAMLP model, in which the historical states are weighted as the updated teacher model to constrain the consistent features.
    \item We propose a strong baseline that exploits the MLP-like model for DG tasks for the first time, achieving the state-of-the-art performance on three benchmarks including PACS, Office-Home and Digits-DG.
\end{itemize}

\begin{figure*}[!t]
  \centering
  \includegraphics[width=.99\textwidth]{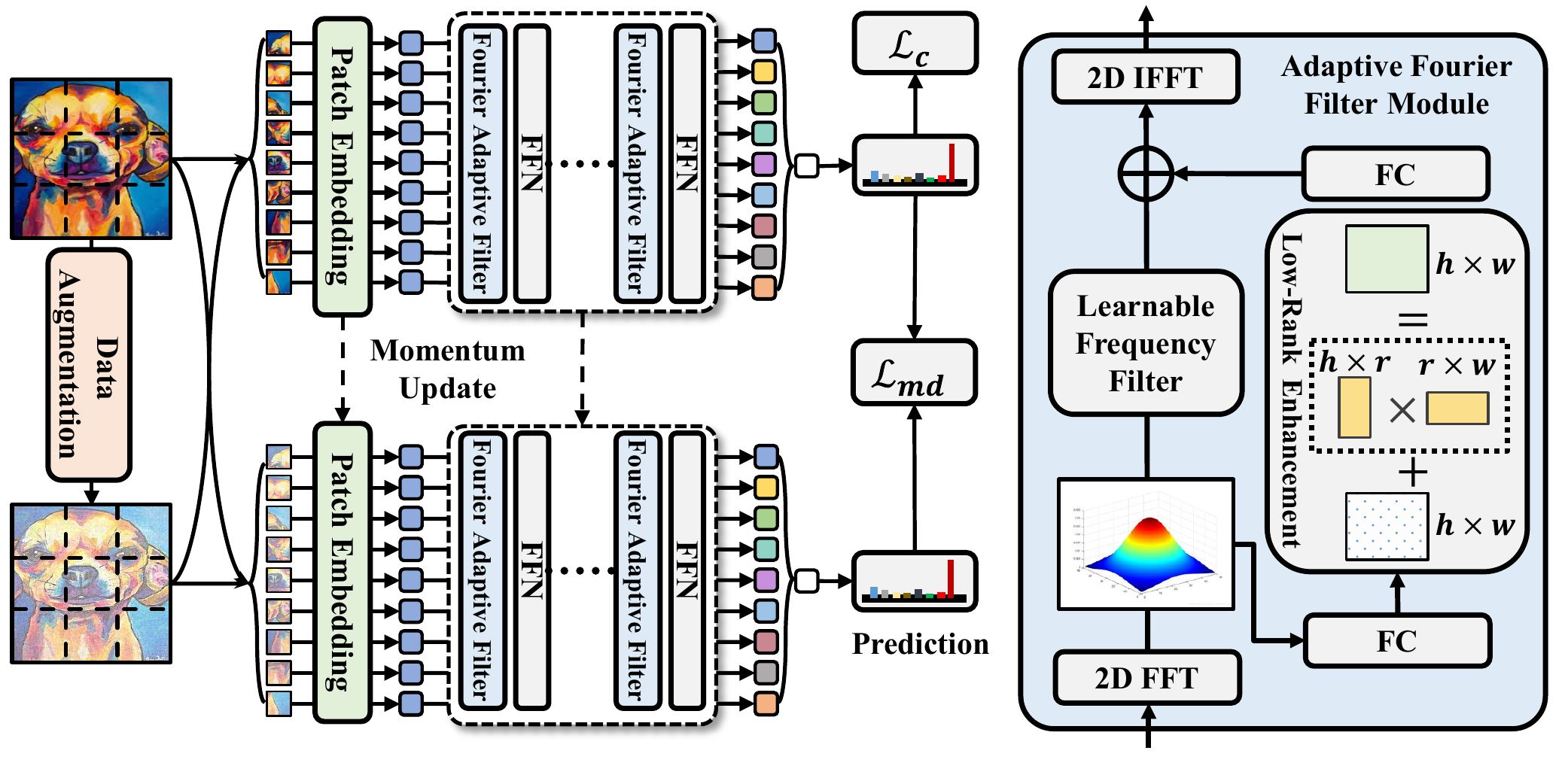}
  \caption{Illustration of the proposed frequency-aware MLP architecture. Specifically, an adaptive Fourier filter (AFF) module is proposed to plug into the MLP-like model. Within this module, a learnable frequency filter (LFF) is utilized to adjust the amplitude distribution by optimizing both the real and imaginary parts. Meanwhile, a low-rank enhancement (LRE) module is further proposed to rectify the filtered features by adding the low-frequency components from SVD decomposition.
  In addition, input images are transformed by the data transformations to prepared data for the distillation loss. 
  }
  \label{Fig_method}
  \end{figure*}
\section{Related Work}

\noindent\textbf{Domain Generalization.}
Domain generalization (DG) targets to generalize the model to  unseen domains with multiple disjoint domains provided during training. Many approaches focus on extracting the domain-invariant features and align the distribution of different domains to address the DG problem. For example, \cite{li2018deep} proposes a conditional invariant adversarial network to guarantee the domain-invariant property and the Siamese network is introduced in \cite{motiian2017unified} to learn a discriminative embedding space. Later, some meta-based \cite{li2018learning,balaji2018metareg} methods are proposed to introduce a type of regularization into the domain generalization. This type of method synthesizes virtual testing domains to simulate train/test domain shift within
each mini-batch. Data augmentation is also a popular idea to address this problem. Adversarial-based \cite{volpi2018generalizing} and Fourier-based \cite{xu2021fourier} examples are generated to improve the generalization of the models. There are also other methods employing low-rank decomposition~\cite{li2017deeper} or self-supervising jigsaw task \cite{carlucci2019domain} to train the models. Convolutional neural networks dominate the task among all of the above methods while we target to investigate the performance of the MLP-like model for DG in this paper.

\noindent\textbf{MLP-Like Backbones.} Recently, some works~\cite{tolstikhin2021mlp,touvron2021resmlp,liu2021pay,rao2021global,hou2022vision} try to replace the self-attention layer with the fully connected layer for the better trade-off between performance and efficiency on the large-scale datasets. MLP-mixer~\cite{tolstikhin2021mlp} firstly proposes a technically simple architecture solely based on multi-layer perceptrons, which mix the per-location features and spatial features. The experimental results show that MLP-like models are as good as existing SOTA methods including CNNs and transformers~\cite{dosovitskiy2020image,park2022vision}. Following this work, gMLP~\cite{liu2021pay} enhances the spatial interaction with multiplicative gating. ResMLP~\cite{touvron2021resmlp} replaces the batch or channel normalization with the simple affine transformation for better trade-off. Vip~\cite{hou2022vision} separately encodes the feature representations along the height and width dimensions for precise positional information. Furthermore, MLP-like models have also been explored in other vision tasks such as dense prediction~\cite{chen2021cyclemlp} and video recognition~\cite{zhang2021morphmlp}. Orthogonally, this paper is designed to explore the transfer capability and optimization strategies of MLP-like models, especially in domain generalization. We believe this is a must for the MLP-like model to act as a universal backbone. To the best of our knowledge, FAMLP is the first method designed for the domain generation task.

\noindent\textbf{Matrix Decomposition.} Matrix decomposition has been widely adopted in deep networks for different purposes. Most researchers focus on network compression by factorizing the low-rank components, including the softmax layer~\cite{sainath2013low}, the convolution layer~\cite{garipov2016ultimate,wang2018wide} and the embedding layer~\cite{lan2019albert}. Recently, some researchers also explore and introduce the certain properties of the decomposed signals to different tasks. \cite{kanakis2020reparameterizing} decomposes each convolution into a shared part for the subsequent incremental tasks.  \cite{geng2021attention} factorizes the representation to recover a clean signal subspace as the global context, modeling the long-range dependencies. \cite{li2021revisiting} revisits the dynamic convolution via matrix decomposition, mitigating the joint optimization difficulty. In contrast, this paper is designed to explore the low-rank components of frequency features for the augmentation on domain-invariant information.
%\noindent\textbf{Knowledge Distillation.}: Knowledge Distillation-based Consistency regularization (CR) is widely-used in supervised and semi-supervised learning. Laine and Aila~\cite{Laine2017TemporalEF} first introduce a consistency loss between outputs from two differently perturbed models.Tarvainen and Valpola~\cite{tarvainen2017mean} further extend this work by using a momentum-updated teacher to provide better targets for consistency alignment. Miyato \etal~\cite{Miyato2019VirtualAT} and Park \etal~\cite{AAAI1816322} develop different techniques by replacing the stochastic perturbations with adversarial ones. Verma \etal~\cite{ijcai2019-504} also demonstrate that a interpolation consistency with MixUp~\cite{zhang2018mixup} samples could be helpful. Some recent work ~\cite{french2017self, shu2018a, wu2020dual} also use consistency regularization in UDA to improve the performance on target domain. In our work, we design a dual-formed consistency loss to further bias our model on the phase information.  

\section{Frequency-Aware MLP}
We detail the frequency-aware MLP architecture and its important components in this section. First of all, we demonstrate the paradigms of problem setting and standard MLP-like model, which is adopted as our baseline. Then two proposed core components adaptive Fourier filter layer and momentum update strategy are introduced. Finally, we analyze the optimization flow of the overall pipeline.  

\subsection{Problem Description}
Given multiple source domains $D_s = \{D_1, \ldots, D_K\}$ with $n$ labelled samples $\{\left(x_{i},y_{i}\right)\}_{i=1}^{n}$ in $k$-th domain $D_k$, where $n$ denotes the number of sampled data, the goal of DG methods is to utilize these data to train a model that performs well on the unseen target domain. 
Although most existing DG works are mainly built upon convolutional neural networks to learn domain-invariant representations after conditioning on the class label from known multi-source domains, this work turns to fully explore the pure MLP architecture for comprehensively investigating the performance of MLP-like models on the domain generalization task.

\subsection{Standard MLP-Like Model.}
\label{backbone}
Following the architecture of MLP-mixer~\cite{tolstikhin2021mlp}, the standard MLP-like model consists of a per-patch transform layer, MLP layers and a classification head. Specifically, the input image X is firstly split into a grid of S $\times $ S non-overlapping patches $X_{0}$ ($X_{0} \in \mathbb{R}^{S^{2}\times \left ( \frac{H}{S} \times \frac{W}{S}\times 3 \right )  } $), where H and W represents the initial spatial size. Then, each patch is independently projected to the embedding space by a linear layer $W_{l}$,
\begin{equation}
  X_{1} = X_{0}W_{l}, W_{l} \in \mathbb{R}^{\frac{3HW}{S^{2}} \times C}.
\end{equation}
The resulting latent features $X_{1}$ are fed to a sequence of MLP layers $W_{mlp}$, which fuse the per-patch and per-channel information in turn,
\begin{equation}
\begin{aligned}
  X_{i+1} = X_{i}W_{\text{mlp}}^{i}, \\
  X_{N+1} = X_{1}W_{\text{mlp}},
  \end{aligned}
\end{equation}
where N represents the number of MLP layers and i represents the \textit{i-th} MLP layer in the sequence. Finally, the output features are averaged as a d-dimension vector, which is fed to a linear classifier $W_{fc}$ for the predicted label,
\begin{equation}
  Y_{\text{pred}} = (\text{Avg}(X_{N+1}))W_{\text{fc}}.
\end{equation}

\textbf{MLP Layer.}
To facilitate the feature interaction during the optimization process, each MLP layer contains two MLP blocks along different dimensions. The input features $X_{i}$ are firstly projected along the patch dimension (\ie, $W_{p}$) in the first block ($\mathbb{R}^{S^{2}} \to \mathbb{R}^{S^{2}}$). To reduce the difficulty of optimization, the initial input is added through the skip connection. Similarly, the middle features $Z_{i}$ are then projected along the channel dimension (\ie, $W_{c}$) in the second block ($\mathbb{R}^{C} \to \mathbb{R}^{C}$). Each MLP block consists of two fully connected layers and an element-wise nonlinearity (GELU \cite{hendrycks2016gaussian}) $\sigma$:
\begin{equation}
\label{eq:mlp}
\begin{aligned}
  Z_{i} = X_{i} + W_{p_{2}}  \sigma  (W_{p_{1}} \text{LayerNorm}(X_{i})), \\
  X_{i+1} = Z_{i} + \sigma  (\text{LayerNorm}(Z_{i}) W_{c_{1}})W_{c_{2}},
\end{aligned}
\end{equation}
where LayerNorm represents the layer normalization \cite{ba2016layer}.

\subsection{Adaptive Fourier Filter Layer}
As the receptive filed of fully connected layer spans a long range and covers global interactions, the extracted features contain extensive domain information, which is reflected in the high-frequency component. To eliminate its negative effect on DG, we add an adaptive Fourier filter layer before each MLP layer. The input features $X_{i}$ are firstly fed to the adaptive Fourier filter layer, eliminating the high-frequency interference. 
\begin{equation}
  X_{i}^{\text{aff}} = X_{i}W_{\text{aff}},
\end{equation}
In this case, equation \ref{eq:mlp} can be rewritten as:
\begin{equation}
  Z_{i} = X_{i}^{\text{aff}} + W_{p_{2}} \sigma (W_{p_{1}}\text{LayerNorm}(X_{i}^{\text{aff}})),
\end{equation}

\textbf{Learnable Frequency Filter.}
To explicitly filter the high-frequency interference in the latent features, we directly transform the spatial feature into the frequency domain. For a input embedding $X_{i}$, its Fourier transformation $\mathcal{F}(X)$ can be formulated as:
\begin{equation}
  \mathcal{F}(X_{i}) (u,v)=\sum_{h=0}^{C-1} \sum_{w=0}^{S^{2}-1} X_{i}(h,w) e^{-j2\pi \left (\frac{h}{C}u+\frac{w}{S^{2}}v    \right ) } 
  = R(X_{i}) + I(X_{i})j,
\end{equation}
\begin{equation}
\begin{aligned}
  A(X_{i}) = \left [R^{2}(X_{i}) + I^{2}(X_{i})  \right ] ^{1/2}, \\
  P(X_{i}) = arctan\left [\frac{I(X_{i})}{R(X_{i})}   \right ],
\end{aligned}
\end{equation}
$R(X_{i})$ and $I(X_{i})$ represent the real and imaginary parts of $\mathcal{F}(X)$. $A(X_{i})$ and $P(X_{i})$ represent the amplitude and phase components in the frequency domain. Existing works~\cite{xu2021fourier,yang2020fda} have proven that the amplitude components is highly related to the domain information, which is influenced by both the real and imaginary parts. To adaptively refine the domain-invariant features, we maintain a learnable frequency filter $W_{filter}$, which is the same size as $R(x_{i})$ and $I(x_{i})$. Different from the small-size filter (\eg, 3$\times$3) in the spatial domain, the frequency filter contains all the sampled frequency values. The frequency features are directly element-wise multiplied by the filter and optimized to adjust the useful amplitude distribution from multiple domains,
\begin{equation}
  Z_{f} = \mathcal{F}(X_{i}) \odot W_{\text{filter}}.
\end{equation}
The filtered features are transformed to the spatial domain through the inverse Fourier transformation for the subsequent operation,
\begin{equation}
%   X_{i}^{\text{aff}} = \mathit{} F(X)^{-1} (Z_{f}),
  X_{i}^{\text{aff}} = \mathit{} \mathcal{F}^{-1}[Z_{f}].
\end{equation}
Both Fourier transformation and the inverse one can be implemented by the FFT algorithm \cite{nussbaumer1981fast}.

\textbf{Low-Rank Enhancement Module.}
To further enhance the maintanence of domain-invariant features, we extract low-frequency components from the perspective of matrix decomposition. A input embedding can be seen as a static kernel $\hat{X}$ and some noise information E, and the latter is sensitive to the variant such as domain shifts,
\begin{equation}
  X_{i} = \hat{X}_{i} + E = DC + E,
\end{equation}
where D and C represent the decomposed matrices, respectively. We utilize the SVD decomposition for the low-rank component in the frequency feature, 
\begin{equation}
  \min_{D, C} L_{re}(X_{i}, DC) + R_{1}(D) + R_{2}(C),
\end{equation}
where $L_{re}$ represents the reconstruction loss, $R_{1}$ and $R_{2}$ are the regularization terms. It is noted that the whole process is non-parameter, so we denote it as $M_{SVD}$, which is distinguished from the learnable operator $W$. To reduce the complexity of $M_{SVD}$, we utilize two linear transformation layers (\ie, $W_{down}$ and $W_{up}$) to map the features to different embedding spaces. Finally the compact features are added to the filtered features for further augmentation,
\begin{equation}
Z_{i}^{\text{aff}} = Z_{f} + W_{\text{up}} M_{\text{SVD}} (W_{\text{down}}\mathcal{F}(X_{i})),
\end{equation}
\begin{equation}
%   X_{i}^{\text{aff}} = \mathit{} F(X)^{-1} (Z_{i}^{\text{aff}}).
  X_{i}^{\text{aff}} = \mathcal{F}^{-1}[Z_{i}^{\text{aff}}].
\end{equation}

\subsection{Momentum Update Strategy}
To enhance the generalization of MLP from the perspective of overall optimization, we adopt the momentum update strategy to the standard full supervised paradigm. Here we denote $W_{\theta _{t}}$ and $W_{\theta _{0\to t}}$ as the all the optimized parameters of student and teacher models at different time state t. We update the teacher model based on the historical state of teacher model and current state of student model for distillation. That is,
\begin{equation}
  W_{\theta _{0\to t}} = \eta W_{\theta _{0\to t-1}} + (1-\eta) W_{\theta _{t}},
\end{equation}
where $\eta$ represents the momentum weight. It is noted that the teacher model can be seen as the weighted summation of the student models, their outputs should be similar. So we constrain the optimized model to be consistent to the teacher one. To further improve the generalization, we adopt data augmentation for the input of the teacher model. It can be seen in the experimental part that these augmentation strategies are also beneficial for the DG problem,  
\begin{equation}
  \tilde{X} = \text{DataAug}(X),
\end{equation} 
\begin{equation}
\begin{aligned}
  \ell _{\text{md}} &= L_{\text{md}}(XW_{\theta_{t}}, \tilde{X}W_{\theta_{0\to t}}) \\&= D_{\text{KL}}(s(W_{fc}(XW_{\theta_{t}})/\tau_{md}) || s(W^{t}_{fc}(\tilde{X}W_{\theta_{0\to t}})/\tau_{md})),
\end{aligned}
\end{equation}
where DataAug represents the Fourier-based data augmentation \cite{xu2021fourier} together with the standard augmentation protocols, $D_{\text{KL}}$ represents Kullback-Leibler divergence, $W^{t}_{fc}$ denotes to the classification head of teacher model, $\tau_{md}$ is the temperature and $s(\cdot)$ refers to the softmax operation. 
% The Fourier-based data augmentation \cite{xu2021fourier} together with the standard augmentation protocol, \ie, \ random resized cropping, horizontal flipping and color jittering are applied in our experiments.

\subsection{Optimization}
Combining all above loss functions together, we can get our full objective when given the input image-target pair (X, Y):
\begin{equation}
\begin{aligned}
  \ell _{\text{all}} & = \ell _{\text{c}}+ \lambda_{md} \ell _{\text{md}} \\
              & = L_{\text{c}} (XW_{\theta_{t}}, Y) + \lambda_{md} L_{\text{md}}(XW_{\theta_{t}}, \tilde{X}W_{\theta_{0\to t}}),
\end{aligned}
\end{equation}
where $L_{\text{c}}$ represents the cross-entropy loss, and $\lambda_{md}$ controls the trade-off between the classification and the distillation loss.

\begin{table}[!t]
  \small
  \centering
  \caption{Leave-one-domain-out results on PACS. The best and second-best results are bolded and underlined respectively. }
    \setlength{\tabcolsep}{1.0mm}
    {\begin{tabular}{l|cccc|c}
    \toprule
    Methods & Art & Cartoon & Photo & Sketch & Avg. \\
    \midrule
    \multicolumn{6}{c}{\textit{ResNet-18}} \\
    \midrule
    DeepAll                             & 77.63 & 76.77 & 95.85 & 69.50 & 79.94 \\
    MetaReg~\cite{balaji2018metareg}    & 83.70 & 77.20 & 95.50 & 70.30 & 81.70 \\
    JiGen~\cite{carlucci2019domain}     & 79.42 & 75.25 & 96.03 & 71.35 & 80.51 \\
    Epi-FCR~\cite{li2019episodic}       & 82.10 & 77.00 & 93.90 & 73.00 & 81.50 \\
    MMLD~\cite{matsuura2020domain}      & 81.28 & 77.16 & 96.09 & 72.29 & 81.83 \\
    DDAIG~\cite{zhou2020deep}           & 84.20 & 78.10 & 95.30 & 74.70 & 83.10 \\
    CSD~\cite{piratla2020efficient}     & 78.90 & 75.80 & 94.10 & 76.70 & 81.40 \\
    InfoDrop~\cite{shi2020informative}  & 80.27 & 76.54 & 96.11 & 76.38 & 82.33 \\
    MASF~\cite{dou2019domain}           & 80.29 & 77.17 & 94.99 & 71.69 & 81.04 \\
    L2A-OT~\cite{zhou2020learning}      & 83.30 & 78.20 & 96.20 & 73.60 & 82.80 \\
    EISNet~\cite{wang2020learning}      & 81.89 & 76.44 & 95.93 & 74.33 & 82.15 \\
    RSC~\cite{huang2020self}            & 83.43 & 80.31 & 95.99 & 80.85 & 85.15 \\
    FACT~\cite{xu2021fourier}           & 85.37 & 78.38 & 95.15 & 79.15 & 84.51 \\
    ATSRL~\cite{yang2021adversarial}    & \underline{85.80} & 80.70 & \underline{97.30} & 77.30 & 85.30 \\
    DIRT-GAN~\cite{nguyen2021domain}    & 82.56 & 76.37 & 95.65 & 79.89 & 83.62 \\
    FSDCL~\cite{jeon2021feature}        & 85.30 & \underline{81.31} & 95.63 & \underline{81.19} & \underline{85.86} \\
    \midrule
    \multicolumn{6}{c}{\textit{MLP-S}} \\
    \midrule
    % \textit{Baseline} & \textbf{} & \textbf{} & \textbf{} &  &	\textbf{} \\
    \textit{Our FAMLP-S}                       & \textbf{92.06} & \textbf{82.49} & \textbf{98.10} & 	\textbf{84.09} &	\textbf{89.19} \\  	 	 	 	
    \midrule
    \multicolumn{6}{c}{\textit{ResNet-50}} \\
    \midrule
    DeepAll                             & 84.94 & 76.98 & 97.64 & 76.75 & 84.08 \\
    MetaReg~\cite{balaji2018metareg}    & 87.20 & 79.20 & 97.60 & 70.30 & 83.60 \\
    MASF~\cite{dou2019domain}           & 82.89 & 80.49 & 95.01 & 72.29 & 82.67 \\
    EISNet~\cite{wang2020learning}      & 86.64 & 81.53 & 97.11 & 78.07 & 85.84 \\
    RSC~\cite{huang2020self}            & 87.89 & 82.16 & 97.92 & 83.35 & 87.83 \\
    FACT~\cite{xu2021fourier}           & 89.63 & 81.77 & 96.75 & \underline{84.46} & \underline{88.15} \\
    ATSRL~\cite{yang2021adversarial}    & \underline{90.00} & 83.50 & \underline{98.90} & 80.00 & 88.10 \\
    MBDG~\cite{robey2021model}          & 80.60 & 79.30 & 97.00 & \textbf{85.20} & 85.60 \\
    FSDCL~\cite{jeon2021feature}        & 88.48 &  \underline{83.83} & 96.59 & 82.92 & 87.96 \\
    SWAD~\cite{cha2021swad}             & 89.30 & 83.40 & 97.30 & 82.50 & 88.10 \\
    \midrule
    \multicolumn{6}{c}{\textit{MLP-B}} \\
    \midrule
    % \textit{Baseline} & {85.00} & {77.86} & {94.43} & 79.33 &{80.75} \\
    \textit{Our FAMLP-B}                       & \textbf{92.63} & \textbf{87.03} & \textbf{98.14} & 82.69 &	\textbf{90.12} \\
    \bottomrule
    \end{tabular}}
  \label{pacs}
\end{table}

\begin{table}[!t]
  \small
  \centering
  \caption{Leave-one-domain-out results on OfficeHome. The best and second-best results are bolded and underlined respectively.}
    \setlength{\tabcolsep}{1.0mm}{\begin{tabular}{l|cccc|c}
    \toprule
    Methods & Art & Clipart & Product & Real & Avg. \\
    \midrule
    \midrule
    \multicolumn{6}{c}{\textit{ResNet-18}} \\
    \midrule
    DeepAll                             & 57.88 & 52.72 & 73.50 & 74.80 & 64.72 \\
    CCSA~\cite{motiian2017unified}      & 59.90 & 49.90 & 74.10 & 75.70 & 64.90 \\
    MMD~\cite{li2018domain}             & 56.50 & 47.30 & 72.10 & 74.80 & 62.70 \\
    CG~\cite{shankar2018generalizing}   & 58.40 & 49.40 & 73.90 & 75.80 & 64.40 \\
    DDAIG~\cite{zhou2020deep}           & 59.20 & 52.30 & 74.60 & 76.00 & 65.50 \\
    L2A-OT~\cite{zhou2020learning}      & 60.60 & 50.10 & 74.80 & 77.00 & 65.60 \\
    Jigen~\cite{carlucci2019domain}     & 53.04 & 47.51 & 71.47 & 72.79 & 61.20 \\
    RSC~\cite{huang2020self}            & 58.42 & 47.90 & 71.63 & 74.54 & 63.12 \\
    FACT~\cite{xu2021fourier}           & 60.34 & \underline{54.85} & 74.48 & 76.55 & 66.56 \\ 
    ATSRL~\cite{yang2021adversarial}    & \underline{60.70} & 52.90 & \underline{75.80} & \underline{77.20} & \underline{66.70} \\
    FSDCL~\cite{jeon2021feature}        & 60.24 & 53.54 & 74.36 & 76.66 & 66.20 \\
    \midrule
    \multicolumn{6}{c}{\textit{MLP-S}} \\
    \midrule
    % \textit{Baseline} & \textbf{} & \textbf{} & \textbf{} &  &	\textbf{} \\
    \textit{Our FAMLP-S}                       & \textbf{69.34} & \textbf{62.61} & \textbf{79.82} & \textbf{82.00} & \textbf{73.44} \\

    \midrule
    \multicolumn{6}{c}{\textit{ResNet-50}} \\
    \midrule
    Fishr~\cite{rame2021fishr}          & 63.40 & 54.20 & 76.40 & 78.50 & 68.20 \\
    SWAD~\cite{cha2021swad}             & 66.10 & 57.70 & 78.40 & 80.20 & 70.60 \\
    ATSRL~\cite{yang2021adversarial}    & \underline{69.30} & \underline{60.10} & \underline{81.50} & \underline{82.10} & \underline{73.30} \\
    \midrule
    \multicolumn{6}{c}{\textit{MLP-B}} \\
    \midrule
    % \textit{Baseline} & 63.45&56.31&77.81&79.76&69.33\\
    \textit{Our FAMLP-B}                & \textbf{70.53} & \textbf{64.63} & \textbf{81.32} & \textbf{82.79} & \textbf{74.82} \\ 
    \bottomrule
    \end{tabular}}
    
  %\vspace{-10pt}
  \label{officehome}
\end{table}

\begin{table}[!h]
\caption{Leave-one-domain-out results on Digits-DG.}
  \centering
    \setlength{\tabcolsep}{0.8mm}{\begin{tabular}{l|cccc|c}
    \toprule
    Methods & MNIST & MNIST-M & SVHN & SYN & Avg. \\
    \midrule
    DeepAll~\cite{zhou2020deep} & 95.8 & 58.8 & 61.7 & 78.6 & 73.7 \\
    CCSA~\cite{motiian2017unified} & 95.2 & 58.2 & 65.5 & 79.1 & 74.5 \\
    MMD-AAE~\cite{li2018domain} & 96.5 & 58.4 & 65.0 & 78.4 & 74.6 \\
    CrossGrad~\cite{shankar2018generalizing} & 96.7 & 61.1 & 65.3 & 80.2 & 75.8 \\
    DDAIG~\cite{zhou2020deep} & 96.6 & {64.1} & 68.6 & 81.0 & 77.6 \\
    Jigen~\cite{carlucci2019domain} & 96.5& 61.4 & 63.7 & 74.0 & 73.9 \\
    L2A-OT~\cite{zhou2020learning} & {96.7} & 63.9 & {68.6} & {83.2} & {78.1} \\
    FACT~\cite{xu2021fourier} & {97.9} & {65.6} & {72.4} & {90.3} & {81.5} \\
    \bottomrule   
    \textit{Our FAMLP-S}   &\textbf{98.0} 	& \textbf{83.3}& \textbf{84.1} &	\textbf{96.9}&	\textbf{90.6} \\  	
    \bottomrule    	 	 	
    \end{tabular}}
  \label{digitsdg}
\end{table}

\section{Experiments}

In this section, we demonstrate the superiority of our method on three conventional DG benchmarks and conduct several ablation studies to show the effectiveness of each component.

\subsection{Setup}

\textbf{Datasets. }We conduct the experiments on three benchmark datasets: \textbf{(1) PACS} \cite{li2017deeper} consists of four domains, \ie, \ Art Painting, Cartoon, Photo and Sketch. It totally contains 9991 images of 7 classes. \textbf{(2) Office-Home} \cite{venkateswara2017deep} is also composed of four domains, \ie, \ Art, Clipart, Product and Real World with 15500 images of 65 classes. The model is trained on three domains and tested on the remaining one during experiments. \textbf{(3)Digits-DG}~\cite{zhou2020deep}: a digit recognition benchmark consisted of four classical datasets \textit{MNIST}, \textit{MNIST-M}, \textit{SVHN}, \textit{SYN}. The four datasets mainly differ in font style, background and image quality. We use the original train-validation split in~\cite{zhou2020deep} with 600 images per class per dataset.
% \textbf{(3)Digits-DG}~\cite{zhou2020deep}: a digit recognition benchmark consisted of four classical datasets \textit{MNIST}~\cite{lecun1998gradient}, \textit{MNIST-M}~\cite{ganin2015unsupervised}, \textit{SVHN}~\cite{netzer2011reading}, \text{SYN}~\cite{ganin2015unsupervised}. The four datasets mainly differ in font style, background and image quality. We use the original train-validation split in~\cite{zhou2020deep} with 600 images per class per dataset. 

\noindent\textbf{Implementation Details.} The backbone is detailed in Section \ref{backbone}, which is pretrained on the ImageNet \cite{russakovsky2015imagenet} with $16\times16$ input patch size in all of our experiments. For fair comparison, we adjust the depth and width of FAMLP to ensure comparable model capacity with different CNNs. Finally, we scale the depth by a factor of $\left \{ 1,4 \right \} $ (\ie, MLP-S and MLP-B), corresponding to ResNet-18 and ResNet-50, respectively. The network is trained for 50 epochs with batch size of 16 and weight decay of 5e-4. We use SGD as the optimizer and set the initial learning rate as 0.001 which is decayed by 0.1 at 40 epochs. The Fourier-based data augmentation \cite{xu2021fourier} together with the standard augmentation protocol, \ie, \ random resized cropping, horizontal flipping and color jittering are applied in our experiments. The momentum m for the teacher model is set as 0.9995 and the value of the temperature $\tau_{md}$ is 10 and the $\tau_{r}$ is 1.5. The first weight parameter $\lambda_{md}$ is set to 2 for PACS/Digits-DG, and 200 for OfficeHome and the second one $\lambda_{r}$ is set to 1 for both datasets. We also use a sigmoid ramp-up \cite{xu2021fourier} for the two weights with a length of 5 epochs. The strength of Fourier-based data augmentation is chosen as 1.0 for PACS/Digits-DG, and 0.2 for OfficeHome.

\subsection{Comparison with State-of-the-Art Methods}
\textbf{Domain Generalization.} To better assess the overall performance of our scheme, we compare it with the SOTA methods of domain generalization. As shown in Table \ref{pacs}, \ref{officehome} and Table \ref{digitsdg}, our method achieves average improvement of 3\%, 4\% and 9\% accuracy on PACS, OfficeHome and Digits-DG datasets, respectively. It is worth noting that our model (\ie, FAMLP-S) maintains good generalization even when the number of parameters decreases, achieving 3.33\% and 7.24\% improvement. In PACS, our method improves the FACT \cite{xu2021fourier} with ResNet-50 as the backbone by 1.98\% and achieves best on the art, cartoon, and photo domains except the sketch domain. The possible reason is that the content of the sketch is simpler than other domains, where the global interaction is not very beneficial. For the results of the larger dataset Office-Home, FAMLP outperforms other ResNet-18 and ResNet-50 based methods by a large margin on all the held-out domains, which further illustrates the superiority of our method.

\noindent\textbf{MLP-Like Architecture.}
To demonstrate the generalization performance of our FAMLP architecture, we compare it with the SOTA MLP-like models, including MLP-mixer, gMLP, ResMLP and Vip. As shown in Table \ref{tab:MLP}, our method achieves one point improvement in the smaller model and 6 points improvement in the larger one. This demonstrates the effectiveness of our scheme in assisting the MLP-like models to resist the disturbances caused by domain shifts.
% 86.28 	82.77 	96.71 	78.80 	86.14 

\begin{table}[]
\centering
\small
\caption{Effectiveness of each proposed components on PACS and OfficeHome. `AFF' refers to the adaptive Fourier filter layer; `MUS' refers to the momenta update strategy; `LFF' refers to the learnable frequency filter; `LRE' refers to the low-rank enhancement module.}
\begin{tabular}{c|cc|c|cccc|c}
\toprule
\toprule
\multicolumn{1}{c|}{} & \multicolumn{2}{c|}{AFF} & \multicolumn{1}{c|}{} & \multicolumn{5}{c}{PACS}  \\ %\cline{2-3}
\multicolumn{1}{c|}{\multirow{-2}{*}{Backbone}} & \multicolumn{1}{c}{LFF} & \multicolumn{1}{c|}{LRE} & \multicolumn{1}{c|}{\multirow{-2}{*}{MUS}} & Art & Cartoon & Photo & Sketch & \textbf{Avg.} \\ \midrule
ResNet-50     & $\times$    & $\times$   & $\times$   & 85.45           & 79.44          & 96.77            & 79.33          & \textbf{85.25} \\
MLP-B         & $\times$    & $\times$   & $\times$   & 85.00           & 77.86          & 94.43            & 65.72          & \textbf{80.75} \\
ResNet-50     & \checkmark  & $\times$   & $\times$   & 86.28           & 82.77          & 96.71            & 78.80          & \textbf{86.14} \\
MLP-B         & \checkmark  & $\times$   & $\times$   & 89.75           & 81.83          & 97.66            & 81.93          & \textbf{87.79} \\
MLP-B         & \checkmark  & \checkmark & $\times$   & \textbf{93.36}  & 85.24          & \textbf{98.62}   & 82.03          & \textbf{89.81} \\
MLP-B         & \checkmark  & $\times$   & \checkmark & 90.45           & 82.96          & 98.41            & 82.49          & \textbf{88.58} \\
MLP-B         & \checkmark  & \checkmark & \checkmark & 92.63           & \textbf{87.03} & 98.14            & \textbf{82.69} & \textbf{90.12} \\ \midrule \midrule

\multicolumn{1}{c|}{} & \multicolumn{2}{c|}{AFF} & \multicolumn{1}{c|}{} & \multicolumn{5}{c}{Office-home} \\ %\cline{2-3}
\multicolumn{1}{c|}{\multirow{-2}{*}{Backbone}} & \multicolumn{1}{c}{LFF} & \multicolumn{1}{c|}{LRE} & \multicolumn{1}{c|}{\multirow{-2}{*}{MUS}} & Art & Clipart & Product & Real & \textbf{Avg.} \\ \midrule
ResNet-50     & $\times$    & $\times$   & $\times$   & 64.77           & 60.02          & 78.80            & 78.82          & \textbf{70.60} \\
MLP-B         & $\times$    & $\times$   & $\times$   & 63.45           & 56.31          & 77.81            & 79.76          & \textbf{69.33} \\
ResNet-50     & \checkmark  & $\times$   & $\times$   & 66.63           & 57.78          & 80.15            & 80.81	         & \textbf{71.34} \\
MLP-B         & \checkmark  & $\times$   & $\times$   & 68.31           & 63.00          & 81.60            & 82.65          & \textbf{73.89} \\
MLP-B         & \checkmark  & \checkmark & $\times$   & \textbf{69.39}  & 64.16          & \textbf{81.50}   & 82.95          & \textbf{74.50} \\
MLP-B         & \checkmark  & $\times$   & \checkmark & 68.81           & 64.63          & 81.08            & 81.23          & \textbf{73.93} \\
MLP-B         & \checkmark  & \checkmark & \checkmark & 70.53           & \textbf{64.63} & 81.32            & \textbf{82.79} & \textbf{74.82} \\
\bottomrule
\bottomrule
\end{tabular}
\label{tab:ab}
\end{table}

\begin{table}[]
\centering
\small
\caption{Comparison between different MLP-like models on PACS and OfficeHome.}
\begin{tabular}{l|c|ccccc}
\toprule
\toprule
\multicolumn{1}{l|}{}   &   & \multicolumn{5}{c}{PACS}              \\
\multicolumn{1}{c|}{\multirow{-2}{*}{Method}} & \multirow{-2}{*}{Para.} & Art  & Cartoon        & Photo& Sketch         & \textbf{Avg.}  \\
\midrule
gMLP-S~\cite{liu2021pay}   & 20& 86.72  & 80.80 & 97.54& 72.13  & 84.23  \\
Vip-S~\cite{hou2022vision}    & 25& 87.35  & 85.96 & \textbf{98.68} & 80.20  & 88.05  \\
ResMLP-S~\cite{touvron2021resmlp} & 40& 85.50  & 78.63 & 97.07& 72.64  & 83.46 \\
MLP-B~\cite{tolstikhin2021mlp}    & 59& 85.00  & 77.86 & 94.43& 65.72  & 80.75  \\ \midrule
Our FAMLP-S    & 25& 92.06  & 82.49 &	98.10& 84.09  & 89.19  \\ 
Our FAMLP-B    & 44& \textbf{92.63} & \textbf{87.03} & 98.14  & \textbf{82.69} & \textbf{90.12}  \\
\midrule
\midrule
\multicolumn{1}{l|}{}   &   & \multicolumn{5}{c}{Office-Home}        \\
\multicolumn{1}{c|}{\multirow{-2}{*}{Method}} & \multirow{-2}{*}{Para.} & Art     & Clipart & Product & Real      & \textbf{Avg.}         \\
\midrule
gMLP-S~\cite{liu2021pay}   & 20 & 64.81 & 58.33 & 75.78 & 79.3           & 69.56 \\
Vip-S~\cite{hou2022vision}    & 25 & 69.55 & 61.51 & 79.34 & \textbf{83.11} & 73.38 \\
ResMLP-S~\cite{touvron2021resmlp} & 40 & 62.42 & 51.94 & 75.40 & 77.21          & 66.74 \\
MLP-B~\cite{tolstikhin2021mlp}    & 59 & 63.45 & 56.31 & 77.81 & 79.76          & 69.33 \\
\midrule
Our FAMLP-S    & 25 & 69.34 & 62.61 & 79.82 & 82.00          & 73.44 \\
Our FAMLP-B    & 44 & \textbf{70.53}& \textbf{64.63}& \textbf{81.32}& 82.79  &\textbf{74.82} \\
\bottomrule
\bottomrule
\end{tabular}
\label{tab:MLP}
\end{table}

\subsection{Ablation Study}
We conduct ablation studies to show the effectiveness of each component in our FAMLP architecture in Table \ref{tab:ab}. The performance of our scheme is mainly attributed to three prominent components: LFF layer, LRE module and MUS. To clarify the function of learnable frequency filter in MLP-like model, we add the LFF layer to both the ResNet-50 and MLP-B model for comparison. It can be seen that the generalization performance of ResNet is initially better than that of MLP, but MLP overtakes ResNet after adding the LFF layer. As analyzed earlier, since the MLP-like model covers global interactions, it contains a large amount of domain information. Although the LFF layer brings gain to the CNNs as well, MLP-like model can benefit more from the frequency operation, which proves the effectiveness of frequency filtering for MLP generalization. Then we add the LRE module and MUS to the Fourier-based MLP-like model separately. We can see that these two components improve the baseline of 1.32\% and 0.42\% on average, which demonstrate the effectiveness of the two components. And the model performs best after the combination of both components, which further shows that the two components act in different ways and can assist each other.

\subsection{Analysis}
\begin{figure}[tb]
  \centering
  \includegraphics[width=.999\textwidth]{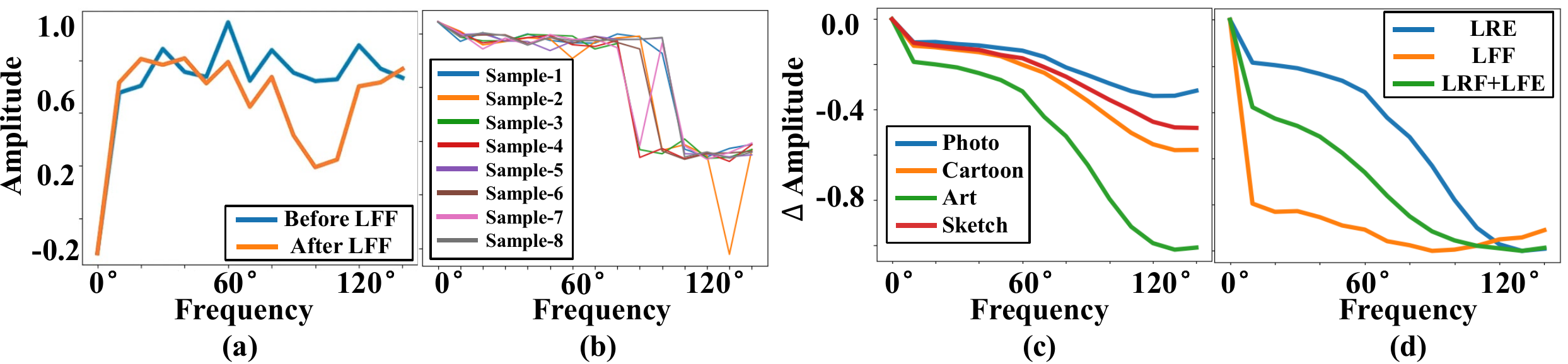}
  \caption{The Fourier analysis of our FAMLP on PACS dataset. (a) The amplitudes before/after the learnable frequency filter; (b) The amplitudes of different samples on sketch domain; (c) $\Delta$ amplitude is the difference between the amplitudes before the learnable frequency filter and after it on different domains; (d) $\Delta$ amplitude is the difference between the amplitudes before/after the different components (\ie, Learnable Frequency  Filter (LFF) and Low-Rank Enhancement module (LRE)). 
  }
%   \vspace{-10pt}
  \label{fig:visual}
\end{figure}

\noindent\textbf{Effectiveness of Learnable Frequency Filter.}
To better demonstrate the role of the learnable frequency filter during optimization, we show the visualization results in the frequency domain. As shown in Fig. \ref{fig:visual} (a), the high-frequency components are obviously suppressed by our Fourier filter. Due to this property, the domain-specific features are greatly filtered out, which improves the generalization of the optimized features. In Fig. \ref{fig:visual} (b) and (c), the vertical coordinate represents the amplitude attenuation of different frequency components before and after adopting the frequency filter. It can be seen that the suppression frequency characteristics are consistent within the domain (\ie, 8 different samples in Fig. \ref{fig:visual} (c)) and different between the domains ((\ie, 4 different domains in Fig. \ref{fig:visual} (b))) owing to our learnable frequency filter. It is the learnability of the frequency filtering kernel that allows the network to adjust adaptively to the domain characteristics of the input, thus enhancing the overall generalization performance of our MLP-like model. 
\noindent\textbf{Effectiveness of Low-Rank Enhancement Module.} To further demonstrate the specific role of low-rank enhancement module, we decompose the visualization results for different layers. As shown in Fig. \ref{fig:visual} (d), only learnable frequency filter tends to oversuppress high frequencies, leading to even some important low-frequency information being lost. To ensure the integrity of the features, the low-rank enhancement module is introduced to augment the low-frequency components. The resulting adaptive Fourier filter layer significantly facilitate the preservation of low-frequency information, thus ensuring the discrimination of the extracted features.

\begin{figure*}[h]
\centering
\includegraphics[scale=0.62]{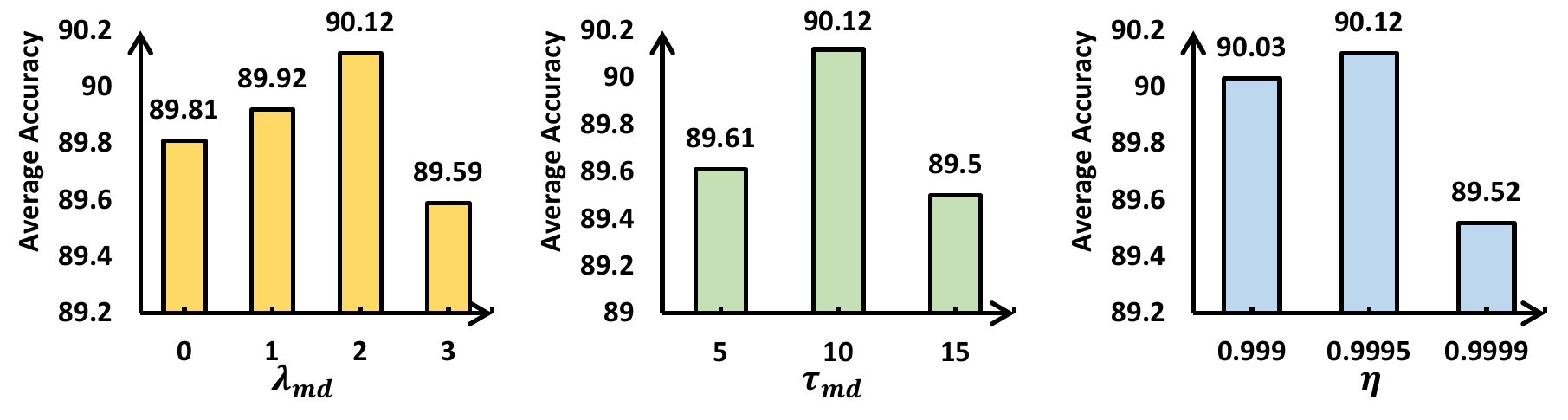}
\caption{Ablation studies of hyper-parameter on PACS dataset. Comparison of average performance when varying the momentum weight $\eta$, weighting factor $\lambda_{md}$ and temperature $\tau_{md}$. }
\label{ab}
\end{figure*}

\noindent\textbf{Effectiveness of Hyper-Parameter.} 
In this subsection, we conduct a series of analysis studies to show how the average accuracy varies as a function of the hyperparameters. The basic values of the $\lambda_{md}$, $\tau_{md}$ and $\eta $ are set to \{2, 10, 0.9995\}. We vary the value of each hyper-parameter and keep the remaining fixed. As shown in Figure \ref{ab}, we can see that the performance indeed changes with the parameters. However, the margin of change is relatively small which means that our method is insensitive to the hyperparameters.

\section{Conclusion}
In this paper, a novel frequency-aware MLP architecture (FAMLP) is presented for the domain generalization task. An adaptive Fourier filter layer is especially designed to be embedded before each MLP layer, augmenting the domain-invariant feature descriptor for label prediction. Specifically, a learnable frequency filter is firstly utilized to adaptively filter out the high-frequency components by considering both the real and imaginary parts of the transformed frequency features. Then, A low-rank enhancement module is further proposed to rectify the filtered features by fusing the low-frequency components from SVD decomposition. In particular, a momentum update strategy is proposed to stabilize the optimization to parameters and input fluctuations by output distillation with the weighted historical model. Experimental results show that our architecture is superior in both performance and adaptability to the state-of-the-art methods, especially in the smaller model.

% \noindent\textbf{Broader impacts.} Domain generalization task has no access to the training data of target domain, which is beneficial for mitigatin

\clearpage
\bibliographystyle{splncs04}
\bibliography{egbib}

\begin{thebibliography}{10}
\providecommand{\url}[1]{\texttt{#1}}
\providecommand{\urlprefix}{URL }
\providecommand{\doi}[1]{https://doi.org/#1}

\bibitem{ba2016layer}
Ba, J.L., Kiros, J.R., Hinton, G.E.: Layer normalization. arXiv preprint
  arXiv:1607.06450  (2016)

\bibitem{balaji2018metareg}
Balaji, Y., Sankaranarayanan, S., Chellappa, R.: Metareg: Towards domain
  generalization using meta-regularization. NeurIPS  \textbf{31},  998--1008
  (2018)

\bibitem{carlucci2019domain}
Carlucci, F.M., D'Innocente, A., Bucci, S., Caputo, B., Tommasi, T.: Domain
  generalization by solving jigsaw puzzles. In: CVPR. pp. 2229--2238 (2019)

\bibitem{cha2021swad}
Cha, J., Chun, S., Lee, K., Cho, H.C., Park, S., Lee, Y., Park, S.: Swad:
  Domain generalization by seeking flat minima. arXiv  (2021)

\bibitem{chen2021cyclemlp}
Chen, S., Xie, E., Ge, C., Liang, D., Luo, P.: Cyclemlp: A mlp-like
  architecture for dense prediction. arXiv preprint arXiv:2107.10224  (2021)

\bibitem{dosovitskiy2020image}
Dosovitskiy, A., Beyer, L., Kolesnikov, A., Weissenborn, D., Zhai, X.,
  Unterthiner, T., Dehghani, M., Minderer, M., Heigold, G., Gelly, S., et~al.:
  An image is worth 16x16 words: Transformers for image recognition at scale.
  arXiv preprint arXiv:2010.11929  (2020)

\bibitem{dou2019domain}
Dou, Q., Coelho~de Castro, D., Kamnitsas, K., Glocker, B.: Domain
  generalization via model-agnostic learning of semantic features. NeurIPS
  \textbf{32},  6450--6461 (2019)

\bibitem{garipov2016ultimate}
Garipov, T., Podoprikhin, D., Novikov, A., Vetrov, D.: Ultimate tensorization:
  compressing convolutional and fc layers alike. arXiv preprint
  arXiv:1611.03214  (2016)

\bibitem{geng2021attention}
Geng, Z., Guo, M.H., Chen, H., Li, X., Wei, K., Lin, Z.: Is attention better
  than matrix decomposition? arXiv preprint arXiv:2109.04553  (2021)

\bibitem{hendrycks2016gaussian}
Hendrycks, D., Gimpel, K.: Gaussian error linear units (gelus). arXiv preprint
  arXiv:1606.08415  (2016)

\bibitem{hou2022vision}
Hou, Q., Jiang, Z., Yuan, L., Cheng, M.M., Yan, S., Feng, J.: Vision
  permutator: A permutable mlp-like architecture for visual recognition. IEEE
  Transactions on Pattern Analysis and Machine Intelligence  (2022)

\bibitem{huang2020self}
Huang, Z., Wang, H., Xing, E.P., Huang, D.: Self-challenging improves
  cross-domain generalization. In: ECCV. pp. 124--140. Springer (2020)

\bibitem{jeon2021feature}
Jeon, S., Hong, K., Lee, P., Lee, J., Byun, H.: Feature stylization and
  domain-aware contrastive learning for domain generalization. In: ACM MM. pp.
  22--31 (2021)

\bibitem{kanakis2020reparameterizing}
Kanakis, M., Bruggemann, D., Saha, S., Georgoulis, S., Obukhov, A., Gool, L.V.:
  Reparameterizing convolutions for incremental multi-task learning without
  task interference. In: European Conference on Computer Vision. pp. 689--707.
  Springer (2020)

\bibitem{lan2019albert}
Lan, Z., Chen, M., Goodman, S., Gimpel, K., Sharma, P., Soricut, R.: Albert: A
  lite bert for self-supervised learning of language representations. arXiv
  preprint arXiv:1909.11942  (2019)

\bibitem{li2017deeper}
Li, D., Yang, Y., Song, Y.Z., Hospedales, T.M.: Deeper, broader and artier
  domain generalization. In: ICCV. pp. 5542--5550 (2017)

\bibitem{li2018learning}
Li, D., Yang, Y., Song, Y.Z., Hospedales, T.M.: Learning to generalize:
  Meta-learning for domain generalization. In: AAAI (2018)

\bibitem{li2019episodic}
Li, D., Zhang, J., Yang, Y., Liu, C., Song, Y.Z., Hospedales, T.M.: Episodic
  training for domain generalization. In: CVPR. pp. 1446--1455 (2019)

\bibitem{li2018domain}
Li, H., Pan, S.J., Wang, S., Kot, A.C.: Domain generalization with adversarial
  feature learning. In: CVPR. pp. 5400--5409 (2018)

\bibitem{li2018deep}
Li, Y., Tian, X., Gong, M., Liu, Y., Liu, T., Zhang, K., Tao, D.: Deep domain
  generalization via conditional invariant adversarial networks. In: ECCV. pp.
  624--639 (2018)

\bibitem{li2021revisiting}
Li, Y., Chen, Y., Dai, X., Liu, M., Chen, D., Yu, Y., Yuan, L., Liu, Z., Chen,
  M., Vasconcelos, N.: Revisiting dynamic convolution via matrix decomposition.
  arXiv preprint arXiv:2103.08756  (2021)

\bibitem{liu2021pay}
Liu, H., Dai, Z., So, D., Le, Q.: Pay attention to mlps. Advances in Neural
  Information Processing Systems  \textbf{34} (2021)

\bibitem{matsuura2020domain}
Matsuura, T., Harada, T.: Domain generalization using a mixture of multiple
  latent domains. In: AAAI. vol.~34, pp. 11749--11756 (2020)

\bibitem{motiian2017unified}
Motiian, S., Piccirilli, M., Adjeroh, D.A., Doretto, G.: Unified deep
  supervised domain adaptation and generalization. In: ICCV. pp. 5715--5725
  (2017)

\bibitem{nguyen2021domain}
Nguyen, A.T., Tran, T., Gal, Y., Baydin, A.G.: Domain invariant representation
  learning with domain density transformations. arXiv  (2021)

\bibitem{nussbaumer1981fast}
Nussbaumer, H.J.: The fast fourier transform. In: Fast Fourier Transform and
  Convolution Algorithms, pp. 80--111. Springer (1981)

\bibitem{park2022vision}
Park, N., Kim, S.: How do vision transformers work? arXiv preprint
  arXiv:2202.06709  (2022)

\bibitem{piratla2020efficient}
Piratla, V., Netrapalli, P., Sarawagi, S.: Efficient domain generalization via
  common-specific low-rank decomposition. In: ICML. pp. 7728--7738. PMLR (2020)

\bibitem{rame2021fishr}
Rame, A., Dancette, C., Cord, M.: Fishr: Invariant gradient variances for
  out-of-distribution generalization. arXiv  (2021)

\bibitem{rao2021global}
Rao, Y., Zhao, W., Zhu, Z., Lu, J., Zhou, J.: Global filter networks for image
  classification. Advances in Neural Information Processing Systems
  \textbf{34} (2021)

\bibitem{robey2021model}
Robey, A., Pappas, G.J., Hassani, H.: Model-based domain generalization. arXiv
  (2021)

\bibitem{russakovsky2015imagenet}
Russakovsky, O., Deng, J., Su, H., Krause, J., Satheesh, S., Ma, S., Huang, Z.,
  Karpathy, A., Khosla, A., Bernstein, M., et~al.: Imagenet large scale visual
  recognition challenge. IJCV  \textbf{115}(3),  211--252 (2015)

\bibitem{sainath2013low}
Sainath, T.N., Kingsbury, B., Sindhwani, V., Arisoy, E., Ramabhadran, B.:
  Low-rank matrix factorization for deep neural network training with
  high-dimensional output targets. In: 2013 IEEE international conference on
  acoustics, speech and signal processing. pp. 6655--6659. IEEE (2013)

\bibitem{shankar2018generalizing}
Shankar, S., Piratla, V., Chakrabarti, S., Chaudhuri, S., Jyothi, P., Sarawagi,
  S.: Generalizing across domains via cross-gradient training. arXiv  (2018)

\bibitem{shi2020informative}
Shi, B., Zhang, D., Dai, Q., Zhu, Z., Mu, Y., Wang, J.: Informative dropout for
  robust representation learning: A shape-bias perspective. In: ICML. pp.
  8828--8839. PMLR (2020)

\bibitem{tolstikhin2021mlp}
Tolstikhin, I.O., Houlsby, N., Kolesnikov, A., Beyer, L., Zhai, X.,
  Unterthiner, T., Yung, J., Steiner, A., Keysers, D., Uszkoreit, J., et~al.:
  Mlp-mixer: An all-mlp architecture for vision. Advances in Neural Information
  Processing Systems  \textbf{34} (2021)

\bibitem{touvron2021resmlp}
Touvron, H., Bojanowski, P., Caron, M., Cord, M., El-Nouby, A., Grave, E.,
  Izacard, G., Joulin, A., Synnaeve, G., Verbeek, J., et~al.: Resmlp:
  Feedforward networks for image classification with data-efficient training.
  arXiv preprint arXiv:2105.03404  (2021)

\bibitem{venkateswara2017deep}
Venkateswara, H., Eusebio, J., Chakraborty, S., Panchanathan, S.: Deep hashing
  network for unsupervised domain adaptation. In: CVPR. pp. 5018--5027 (2017)

\bibitem{volpi2018generalizing}
Volpi, R., Namkoong, H., Sener, O., Duchi, J., Murino, V., Savarese, S.:
  Generalizing to unseen domains via adversarial data augmentation. arXiv
  (2018)

\bibitem{wang2020learning}
Wang, S., Yu, L., Li, C., Fu, C.W., Heng, P.A.: Learning from extrinsic and
  intrinsic supervisions for domain generalization. In: ECCV. pp. 159--176.
  Springer (2020)

\bibitem{wang2018wide}
Wang, W., Sun, Y., Eriksson, B., Wang, W., Aggarwal, V.: Wide compression:
  Tensor ring nets. In: Proceedings of the IEEE Conference on Computer Vision
  and Pattern Recognition. pp. 9329--9338 (2018)

\bibitem{xu2021fourier}
Xu, Q., Zhang, R., Zhang, Y., Wang, Y., Tian, Q.: A fourier-based framework for
  domain generalization. In: CVPR (2021)

\bibitem{yang2021adversarial}
Yang, F.E., Cheng, Y.C., Shiau, Z.Y., Wang, Y.C.F.: Adversarial teacher-student
  representation learning for domain generalization. NeurIPS  \textbf{34}
  (2021)

\bibitem{yang2020fda}
Yang, Y., Soatto, S.: Fda: Fourier domain adaptation for semantic segmentation.
  In: Proceedings of the IEEE/CVF Conference on Computer Vision and Pattern
  Recognition. pp. 4085--4095 (2020)

\bibitem{zhang2021morphmlp}
Zhang, D.J., Li, K., Chen, Y., Wang, Y., Chandra, S., Qiao, Y., Liu, L., Shou,
  M.Z.: Morphmlp: A self-attention free, mlp-like backbone for image and video.
  arXiv preprint arXiv:2111.12527  (2021)

\bibitem{zhou2020deep}
Zhou, K., Yang, Y., Hospedales, T., Xiang, T.: Deep domain-adversarial image
  generation for domain generalisation. In: AAAI. vol.~34, pp. 13025--13032
  (2020)

\bibitem{zhou2020learning}
Zhou, K., Yang, Y., Hospedales, T., Xiang, T.: Learning to generate novel
  domains for domain generalization. In: ECCV. pp. 561--578. Springer (2020)

\end{thebibliography}
\end{document}